\documentclass[conference,table]{IEEEtran}
\usepackage[utf8]{inputenc}
\usepackage{textgreek}
\IEEEoverridecommandlockouts

\usepackage{bbding}
\usepackage{cite}
\usepackage{amsmath,amssymb,amsfonts}
\usepackage{bm}
\usepackage{textcomp}
\usepackage{caption}
\usepackage{graphicx}
\usepackage{pgfplots}
\pgfplotsset{compat=1.18}
\usepackage{subcaption}
\usepackage{multirow}
\usepackage{adjustbox}
\usepackage{booktabs}
\usepackage{siunitx}
\usepackage{tabularx}
\usepackage{comment}
\usepackage{authblk}
\usepackage{tikz}
\usetikzlibrary{arrows.meta, positioning, shadows}
\usepackage[skip=3pt]{caption}
\usepackage{hyperref}
\hypersetup{
  colorlinks = true,
  citecolor  = blue,
  linkcolor  = blue,
  urlcolor   = blue
}

\usepackage{amssymb}
\usepackage{fancyhdr}
\fancypagestyle{firstpage}
{
    \fancyhead[L]{\footnotesize © 2026 IEEE.  Personal use of this material is permitted.  Permission from IEEE must be obtained for all other uses, in any current or future media, including reprinting/republishing this material for advertising or promotional purposes, creating new collective works, for resale or redistribution to servers or lists, or reuse of any copyrighted component of this work in other works. This paper is accepted at the 2026 IEEE International Conference on Edge Computing and Communications (IEEE EDGE 2026).}
    \fancyhead[R]{}
}
\begin{document}
\title{SENTRY: Statistical Reliability Analysis of Vision Transformers Under Soft Errors
\\}

\author[1]{Pramit Kumar Bhaduri}
\author[1,2]{Mahdi Taheri}
\author[1,3]{Samira Nazari}
\author[2]{Maksim Jenihhin}
\author[1]{Christian Herglotz}
\author[1]{Michael Hübner}

\affil[1]{Brandenburg University of Technology Cottbus-Senftenberg, Germany}
\affil[2]{Tallinn University of Technology, Tallinn, Estonia}
\affil[3]{Zanjan University, Iran}

\maketitle
\thispagestyle{firstpage}

\begin{abstract}
With the growth of Vision Transformers in safety-critical domains like autonomous systems and medical imaging, ensuring their reliability against soft errors is paramount. While ViTs offer state-of-the-art accuracy, their massive parameter counts render exhaustive fault injection campaigns infeasible. To bridge this gap, a statistical fault injection framework is presented, leveraging finite-population sampling theory to provide formal reliability guarantees. It is demonstrated that failure rates are bounded within a 1\% margin at 99\% confidence using only a few thousand samples, regardless of model scale. This methodology achieves up to a $10{,}700\times$ reduction in experimental cost compared to exhaustive approaches, while preserving the ability to localize vulnerabilities across architectural components. Through extensive evaluation of different architectures like ViT-Tiny and ViT-Small, a highly non-uniform reliability landscape is uncovered. It is shown that while only 3\% of FP32 bit-flips result in failure, the vast majority of these events lead to catastrophic accuracy collapse. Specific vulnerabilities are localized to normalization layers and critical exponent bits within the IEEE-754 format, providing a mathematical foundation and actionable insights for the design of hardened, edge-deployed ViT architectures.
\end{abstract}


 
\section{Introduction}
\label{sec:introduction}

Deep neural networks (DNNs) dominate modern computer vision and are increasingly deployed in safety-critical environments, including autonomous driving, medical diagnostics, and aerospace remote sensing~\cite{12, 11, 10, 9}. This trend is accelerated by edge deployment, which operates under stringent power and memory constraints, leaving no margin for hardware redundancy or error-correcting codes. Consequently, a single transient fault corrupting a weight during edge inference can lead to catastrophic system failure. Characterizing which faults matter and quantifying their occurrence probability is therefore a prerequisite for designing robust protection mechanisms. In this landscape, the Vision Transformer (ViT)~\cite{dosovitskiy2021image} has emerged as a powerful alternative to convolutional neural networks (CNNs) by leveraging self-attention mechanisms. However, this architectural shift introduces massive parameter counts; for instance, a standard ViT-Base model contains approximately 86 million parameters, representing over 2.8 billion individual bits that serve as potential soft-error targets ~\cite{dosovitskiy2021image}.

Reliability evaluation is traditionally performed via fault injection~\cite{8, 7, 6, 5, 4}. While conceptually straightforward, exhaustive fault injection is rendered computationally intractable by the sheer scale of modern ViTs. For instance, an exhaustive bit-level campaign for a compact ViT-Tiny model (5.5 million parameters) requires evaluating 176 million fault sites. At a modest execution time of ten seconds per inference, such a campaign demands over 56 years of continuous computation, making exhaustive evaluation practically impossible.

To circumvent this bottleneck, existing studies often rely on heuristic sampling campaigns or bit-error-rate (BER) sweeps~\cite{reagen2018ares}. While useful for observing general degradation trends, these empirical approaches lack statistical confidence guarantees. Alternatively, statistical fault injection (SFI) leverages finite-population sampling theory to estimate failure rates with formal confidence bounds using only a fraction of the population~\cite{leveugle2009statistical}. This methodology was recently extended to deep neural networks through an iterative refinement framework that dynamically adjusts sample sizes~\cite{ruospo2025quantitative}. However, these statistical frameworks have been validated exclusively on CNN architectures (e.g., ResNet, MobileNet). Direct translation to ViTs is obstructed by fundamental architectural differences that ViTs rely heavily on LayerNorm layers and route representations through a continuous, shared residual stream - features entirely absent or structurally distinct in CNNs.

This gap is addressed in this paper by presenting a multi-granularity statistical fault injection framework tailored for Vision Transformers. Unlike previous network-level SFI approaches, the proposed framework operates hierarchically across network, layer, component, and bit-position granularities. By propagating measured failure rates as refined priors between stages, the conservative assumptions of traditional SFI are bypassed, dramatically reducing the required sample size while preserving rigorous confidence bounds. Furthermore, this granular analysis uncovers unique vulnerability patterns inherent to transformer dynamics such as the disproportionate sensitivity of normalization layers and residual connections.

The proposed methodology is evaluated using two distinct configurations, ViT-Tiny trained on EuroSAT and ViT-Small trained on CIFAR-100. By executing approximately 51,000 injection experiments over 80 hours of computation, the framework achieves up to a $10{,}700\times$ reduction in experimental cost compared to exhaustive campaigns, while maintaining a 99\% confidence level and a tight 1\% error bound.

The contributions of this work are summarized as follows:

\begin{enumerate}
    \item A hierarchical multi-granularity statistical fault-injection framework for Vision Transformers that progressively refines reliability estimates across network, layer, component, and bit levels while maintaining formal confidence bounds and reducing the required sampling cost.
    
    \item A scalable reliability analysis methodology that enables cross-architecture evaluation and statistically grounded comparison of Vision Transformer reliability under single-bit fault in floating-point weights.
    
    \item A fine-grained fault-localization analysis that identifies failure-critical numerical representations and architectural components, enabling interpretable reliability characterization at the bit, parameter, and module levels.
\end{enumerate}

The remainder of this paper is organized as follows. Section~\ref{sec:background} reviews the relevant background on Vision Transformers, soft-error fault models, and prior statistical and ViT-focused reliability work. Section~\ref{sec:methodology} describes the fault model, model configurations, statistical campaign structure, and failure classification criteria. Section~\ref{sec:results} reports the experimental results across all granularity levels and both model configurations. Section~\ref{sec:conclusion} concludes the paper.

\section{Background and Related Work}
\label{sec:background}

\subsection{Vision Transformers}
\label{sec:bg_vit} 

The Vision Transformer (ViT)~\cite{dosovitskiy2021image} adapts the transformer architecture to image classification by processing images as sequences of patches. An input image of size $H \times W \times C$ is divided into a grid of non-overlapping patches of size $P \times P \times C$, which are flattened and mapped to a $D$-dimensional embedding via a trainable linear projection. A learnable classification token and positional embeddings are added before the sequence is processed by $L$ transformer encoder blocks.

Each encoder block consists of two primary sub-blocks: multi-head self-attention (MSA) and a two-layer feed-forward Multilayer Perceptron (MLP) with GELU activation~\cite{hendrycks2016gaussian}. Both components employ residual connections and are preceded by Layer Normalization (LN). Crucially, because MSA and MLP write outputs directly into the residual stream, faults in parameters near these write points - such as LN scale/shift, the second MLP layer, and attention output projections - enter the main representational pathway without intervening attenuation, potentially increasing vulnerability.

Standard configurations like ViT-Base/16 utilize approximately 86 million parameters, while smaller variants like ViT-Tiny and ViT-Small utilize 5.5M and 21.7M parameters, respectively. In typical deployments, these parameters are stored in 32-bit floating-point (FP32) format.

\subsection{Soft Errors and the Bit-Flip Fault Model}
\label{sec:bg_soft_errors}

A soft error is a transient fault that corrupts stored data without permanent hardware damage~\cite{baumann2005radiation}, often induced by ionizing radiation or electromagnetic interference. The most common manifestation is a single-event upset (SEU), where one bit in a memory cell or register flips. 

In DNN reliability analysis, the standard model involves injecting random bit-flips into weight parameters. For a 32-bit IEEE-754 value, the impact depends on the bit position: mantissa flips (bits 0--22) cause small perturbations, whereas exponent flips (bits 23--30) can change a weight's magnitude by orders of magnitude, often triggering NaN propagation and catastrophic model failure.

\subsection{Statistical Fault Injection}
\label{sec:bg_sfi}

An exhaustive fault injection campaign evaluates every possible fault and records the resulting system behavior. For a network containing $N$ parameters stored in 32-bit format, this corresponds to $32N$ possible fault locations, which is computationally prohibitive for ViTs. Statistical fault injection (SFI) addresses this by applying finite-population sampling theory to the fault space~\cite{leveugle2009statistical}. The number of samples $n$ required to estimate a failure rate with a margin of error $e$ at confidence level $t$ is:
\begin{equation}
\label{eq:sample_size}
n = \frac{N \cdot t^2 \cdot p \cdot (1 - p)}{e^2 \cdot (N - 1) + t^2 \cdot p \cdot (1 - p)}
\end{equation}
where $p$ is the prior estimate of the failure rate. If $p$ is unknown, a conservative value of $0.5$ is used to maximize the sample size. The actual achieved margin of error $e$ is verified post-campaign as:
\begin{equation}
\label{eq:margin}
e = t \cdot \sqrt{\frac{\hat{p}(1 - \hat{p})}{n}} \cdot \sqrt{\frac{N - n}{N - 1}}
\end{equation}
The term $\sqrt{(N-n)/(N-1)}$ is the finite population correction (FPC), which is vital when analyzing small sub-populations, such as individual layers, where the sampling fraction $n/N$ is non-negligible. For ViT-Tiny ($N \approx 5.5 \text{M}$), a campaign with $e = 0.01$ and 99\% confidence ($t=2.576$) requires only $n = 16{,}538$ experiments-approximately 0.30\% of the fault population.

\subsection{Iterative Refinement of SFI Campaigns}
\label{sec:bg_iterative}

While $p=0.5$ ensures statistical coverage, it overestimates the required samples for neural networks, where failure probabilities typically range between 1\% and 3\%. An iterative refinement strategy is introduced in~\cite{ruospo2025quantitative} to exploit this. The first iteration uses a coarse margin $e_{\text{start}}$ and $p_0=0.5$. Subsequent iterations update the prior as $p_i = \hat{p}_{i-1}$, significantly reducing the sample count. This is further optimized using an adaptive E--P curve:
\begin{equation}
\label{eq:ep_curve}
e_{(i+1)} =
\begin{cases}
    -k\hat{P}_i^2 + k\hat{P}_i + e_{\text{goal}}, & \text{if } \hat{E}_i / 3 > e_{\text{goal}} \\
    e_{\text{goal}}, & \text{otherwise}
\end{cases}
\end{equation}
where $k = 4 \cdot (\hat{E}_i / 3 - e_{\text{goal}})$. This parabolic adjustment accelerates convergence toward the target margin when $\hat{P}$ is near 0 or 1.

Prior validation on CNNs like ResNet-20 and MobileNetV2 demonstrates that this iterative approach reduces injected faults by 66\% to 90\% compared to one-step sampling~\cite{ruospo2025quantitative}. While those studies focus on permanent stuck-at faults and per-image Silent Data Corruption (SDC-1), the current work adapts this sampling theory to transient SEUs in ViTs, defining failure as accuracy degradation over a test subset.

\subsection{Reliability of Vision Transformers}
\label{sec:bg_related}

Vulnerability analysis of ViTs reveals distinct sensitivities compared to CNNs. In~\cite{xue2023soft}, it is observed that ViTs are more resilient in linear components due to softmax stabilization but more sensitive in non-linear operations. Subsequent work in~\cite{he2025fine} identifies that early layers are sensitive to attention projections, while deeper layers rely more heavily on MLP weights. Comparative studies across Swin, DeiT, and ViT architectures~\cite{liao2025analyzing} suggest that ViTs generally degrade faster than CNNs under identical bit error rates (BER).

However, these existing studies primarily rely on BER-sweep experiments with fixed repetition counts. While BER sweeps characterize degradation as a function of error intensity, they do not quantify the probability of failure from a single random SEU with statistical confidence. This paper fills that gap by applying advanced SFI to estimate failure rates and confidence intervals within the specific architectural constraints of Vision Transformers.

\section{Methodology}
\label{sec:methodology}

\begin{figure}[t]
\centering
\includegraphics[width=\columnwidth]{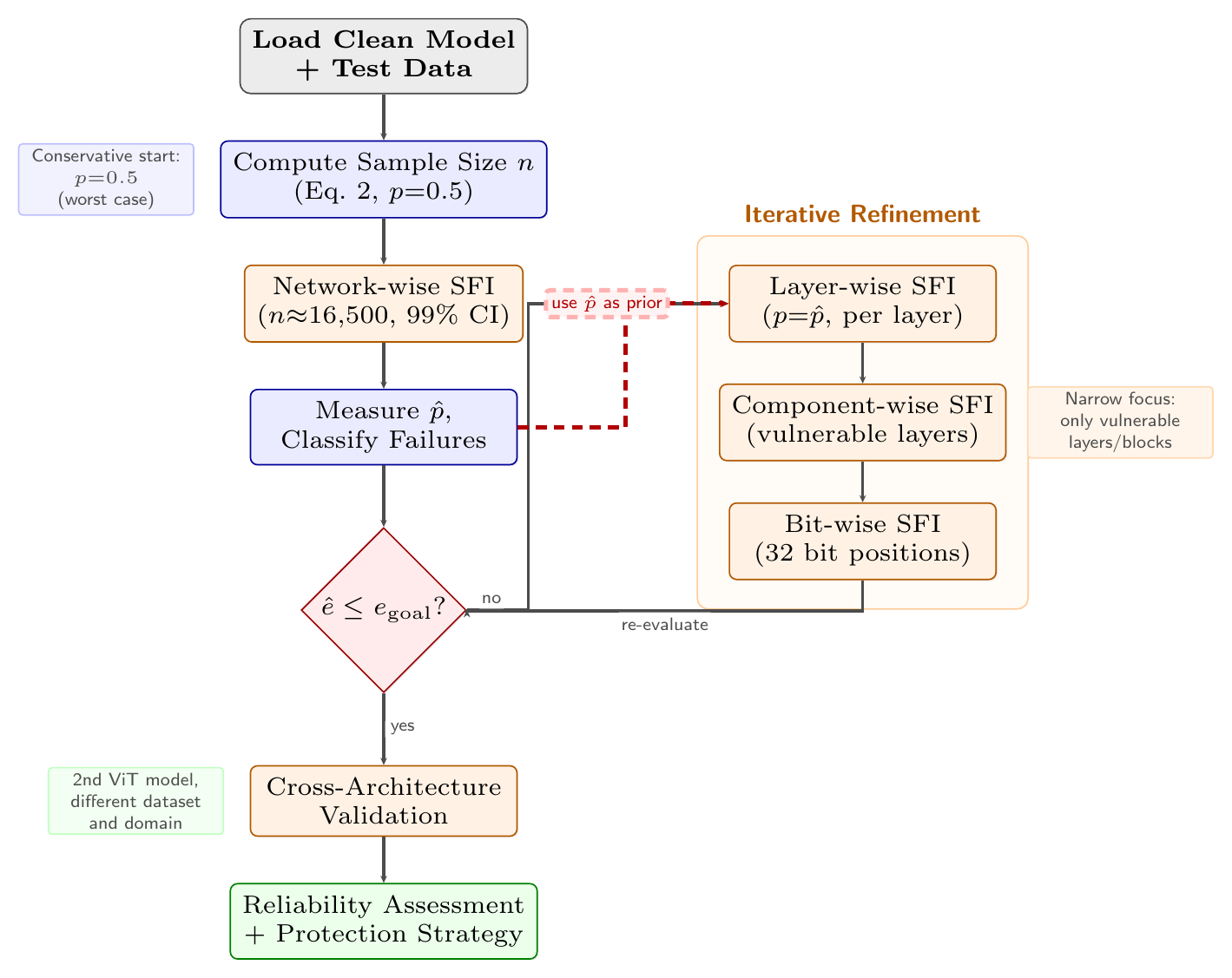}
\caption{Overview of the progressive statistical fault injection workflow, shifting from a conservative network-wise baseline to targeted layer, component, and bit-position granularities.}
\label{fig:methodology}
\end{figure}

This section describes the progressive, multi-granularity statistical fault injection framework designed to characterize the soft-error resilience of Vision Transformers. The framework maps failure probabilities from global network-level behaviors down to local bit-level vulnerabilities. Figure~\ref{fig:methodology} outlines the unified workflow.

\subsection{Mathematical Fault Model}
\label{sec:meth_fault_model}

Single-event upsets are modeled as single transient random bit-flips injected into the model's weight parameters. For any given injection trial, a single trainable parameter tensor is selected uniformly at random from the model's parameter space. A single scalar element $w$ within this tensor is chosen, and a single bit within its 32-bit IEEE-754 floating-point representation is inverted.

Let $w \in \mathbb{R}$ represent the target scalar parameter. The IEEE-754 single-precision format represents $w$ as a tuple of bit fields $(s, e, m)$, where $s \in \{0, 1\}$ is the sign bit, $e \in \{0, 1\}^8$ represents the exponent, and $m \in \{0, 1\}^{23}$ represents the mantissa. The bit-flip operation is formally defined as:
\begin{equation}
\label{eq:bit_flip}
w^* = \text{unpack}\Big(\text{pack}(w) \oplus 2^b\Big)
\end{equation}
where $\text{pack}(\cdot)$ maps the real-valued floating-point number to its equivalent 32-bit unsigned integer representation, $\oplus$ denotes the bitwise exclusive-OR (XOR) operator, $b \in \{0, 1, \dots, 31\}$ is the target bit index, and $\text{unpack}(\cdot)$ casts the modified bit-string back to an IEEE-754 float. This fault model covers all trainable parameter tensors, including projection weights, biases, and LayerNorm scale ($\gamma$) and shift ($\beta$) parameters. Each experiment is executed on a clean parameter state, which is restored immediately after inference.

\subsection{Progressive Multi-Granularity SFI Framework}
\label{sec:meth_campaigns}

To bypass the computational limits of exhaustive testing without losing statistical guarantees, a progressive, multi-granularity SFI framework is defined. Prior techniques applied iterative SFI variations solely to homogeneous populations (e.g., CNN layer activations) using dynamic margin-adjustment curves~\cite{ruospo2025quantitative}. In contrast, the unique structural properties of ViTs-such as LayerNormalization parameters and the unattenuated residual stream-require a multi-stage approach across four logical granularities:

\paragraph{Network-Wise Granularity} The initial campaign evaluates the global vulnerability rate of the entire parameter population $N$. With no prior statistical knowledge, the baseline campaign sets the prior failure probability to its maximum-entropy state, $p_{\text{net}} = 0.5$, under a strict margin of error $e$ and confidence level $t$. This yields a baseline sample size $n_{\text{net}}$ via Equation~\ref{eq:sample_size}, establishing a statistically rigorous global failure rate $\hat{p}_{\text{net}}$.

\paragraph{Layer-Wise Granularity} To identify architectural vulnerabilities, parameters are grouped into $L$ discrete layer populations (e.g., patch embedding, individual encoder blocks, final classification head). Instead of reverting to $p = 0.5$ for each sub-population, the empirical global failure rate $\hat{p}_{\text{net}}$ serves as the prior ($p_{\text{layer}} = \hat{p}_{\text{net}}$) for Equation~\ref{eq:sample_size}. By exploiting the fact that $\hat{p}_{\text{net}} \ll 0.5$, the required sample size per layer $n_{\text{layer}}$ is substantially reduced while preserving statistical confidence.

\paragraph{Component-Wise Granularity} Sub-structures within high-vulnerability encoder blocks are isolated into distinct structural components: query/key/value projections (\texttt{attn\_qkv}), attention output projections (\texttt{attn\_proj}), feed-forward sublayers (\texttt{mlp\_fc1}, \texttt{mlp\_fc2}), and local normalization barriers (\texttt{norm1}, \texttt{norm2}). The localized sample sizes $n_{\text{comp}}$ are calculated using the observed layer-level failure rate $\hat{p}_{\text{layer}}$ as the prior $p_{\text{comp}}$, allowing fine-grained classification of architectural weak points.

\paragraph{Bit-Wise Granularity} The final granularity level isolates numerical sensitivities by restricting bit-flips to specific indices $b \in [0, 31]$ across all parameters. Given the highly non-uniform sensitivity of the IEEE-754 format, the prior $p_{\text{bit}}$ is adjusted according to the targeted bit field:
\begin{equation}
\label{eq:bit_priors}
p_{\text{bit}} = 
\begin{cases}
    p_{\text{mantissa}} \ll 0.5, & b \in [0, 22] \cup \{31\} \\
    p_{\text{exponent}} < 0.5, & b \in [23, 29] \\
    p_{\text{MSB}} > 0.5, & b = 30
\end{cases}
\end{equation}
This targeted initialization ensures that robust bit fields (e.g., mantissa bits) are not over-sampled, while highly sensitive fields (e.g., exponent bits) receive sufficient sampling density to narrow their respective confidence intervals.

\subsection{Failure Classification Criteria}
\label{sec:meth_classification}

Let $A_{\text{clean}}$ be the classification accuracy of the uncorrupted model on a test set, and $A_{\text{fault}}$ be the accuracy under a single bit-flip. The operational consequence of a fault is classified into one of six mutually exclusive categories based on the magnitude of accuracy degradation $\Delta A = A_{\text{clean}} - A_{\text{fault}}$ and numerical stability:

\begin{enumerate}
    \item \textbf{Catastrophic (NaN/Inf):} The corrupted weight triggers non-finite values (NaN or Inf) during inference, causing numerical corruption across the representational pathway.
    \item \textbf{Catastrophic (Low Accuracy):} The accuracy drops below an absolute catastrophic threshold $\tau_{\text{cat}}$, where $\tau_{\text{cat}}$ is scaled to a small multiple of the random-chance baseline $1/C$ for a dataset with $C$ classes.
    \item \textbf{Mission Failure:} The accuracy drops below an operational threshold $\tau_{\text{mission}}$, where $\tau_{\text{cat}} < A_{\text{fault}} < \tau_{\text{mission}}$, representing severe functional degradation that renders the model unusable.
    \item \textbf{SDC Significant:} The model exhibits a noticeable accuracy drop exceeding an acceptable tolerance $\delta_{\text{sig}}$ without crossing the mission failure threshold ($A_{\text{fault}} \ge \tau_{\text{mission}}$ and $\Delta A > \delta_{\text{sig}}$).
    \item \textbf{SDC Minor:} The model suffers a marginal accuracy reduction ($0 < \Delta A \le \delta_{\text{sig}}$).
    \item \textbf{No Effect:} The model accuracy is unaffected or marginally improved ($\Delta A \le 0$).
\end{enumerate}

For formal reliability estimation, the binary failure variable $Y_j \in \{0, 1\}$ for an injection experiment $j$ is defined as:
\begin{equation}
\label{eq:failure_indicator}
Y_j = 
\begin{cases}
    1, & \text{if } \text{Category} \in \{1, 2, 3, 4\} \\
    0, & \text{if } \text{Category} \in \{5, 6\}
\end{cases}
\end{equation}
This classification treats minor SDCs as non-failures to account for test-subset variance while classifying any fault that degrades functional performance beyond acceptable limits as a failure.

\subsection{State-Restoration Optimizations}
\label{sec:meth_implementation}

To maintain high throughput during large-scale campaigns, the SFI framework implements a localized state-restoration algorithm. Rather than re-initializing the entire model architecture or reloading the complete parameter state dictionary $\Theta$ into memory after each trial - an operation that scales poorly with parameter count - the framework tracks the modified tensor address.

Let $\theta_k \in \Theta$ be the specific parameter tensor targeted by the fault injector. Before bit-inversion, a clean copy of $\theta_k$ is retained in local memory. Once inference completes and the failure category is logged, the framework restores only the targeted tensor:
\begin{equation}
\label{eq:restore}
\Theta^* \leftarrow \Theta \setminus \{\theta_k^*\} \cup \{\theta_k\}
\end{equation}
This targeted update minimizes memory bus overhead, making the fault injection framework highly scalable for large transformer configurations.

\section{Experimental Results}
\label{sec:results}

\subsection{Experimental Setup}
\label{sec:setup}

This section describes the experimental configurations, target datasets, training hyperparameters, and the hardware and software environments used to validate the proposed multi-granularity statistical fault injection framework.

The evaluation targets two distinct Vision Transformer scales to verify the architectural generalization of the fault vulnerabilities. Both ViT-Tiny and ViT-Small are lightweight variants suited for resource-constrained deployment, making them representative targets for edge AI reliability assessment.
\begin{itemize}
    \item \textbf{ViT-Tiny Configuration:} Fine-tuned on the EuroSAT satellite imagery dataset~\cite{helber2019eurosat} (10 classes). The network contains $5,526,346$ trainable parameters ($176,843,072$ FP32 bits). The baseline classification model is fine-tuned for one epoch using the AdamW optimizer with a learning rate of $10^{-4}$ and weight decay of $0.05$, achieving a clean top-1 validation accuracy of $97.9\%$ on the full $5,400$-image test set.
    \item \textbf{ViT-Small Configuration:} Fine-tuned on the CIFAR-100~\cite{krizhevsky2009learning} natural image dataset (100 classes). This configuration represents a $3.93\times$ increase in scale, containing $21,704,164$ trainable parameters ($694,533,248$ FP32 bits), featuring $6$ attention heads (compared to $3$ in ViT-Tiny) and an embedding dimension of $384$ (compared to $192$). The model is trained for $10$ epochs using a cosine annealing learning rate schedule, reaching a clean baseline accuracy of $90.99\%$ on the full $10,000$-image test set.
\end{itemize}

To maintain a tractable per-injection evaluation cycle on CPU-based infrastructure, representative subsets of the test datasets are used during fault injection loops:
\begin{itemize}
    \item For the ViT-Tiny/EuroSAT campaign, a fixed subset of $1,000$ test images is evaluated per fault injection.
    \item For the ViT-Small/CIFAR-100 campaign, a single-batch validation subset of $128$ images is evaluated. This is sufficient for reliable pass/fail classification for two reasons: the failure thresholds are far below the 90.99\% clean baseline, and over 92\% of observed failures are catastrophic, meaning accuracy collapses below any threshold regardless of which 128 images are sampled. 
\end{itemize}
The failure classification thresholds $\tau_{\text{cat}}$ (catastrophic low accuracy) and $\tau_{\text{mission}}$ (mission failure) are established relative to the target datasets' class sizes and clean performance baselines. For the 10-class EuroSAT configuration, these thresholds are set to $\tau_{\text{cat\_tiny}} = 20\%$ and $\tau_{\text{mission\_tiny}} = 50\%$. For the 100-class CIFAR-100 configuration, the boundaries are adjusted to $\tau_{\text{cat\_small}} = 10\%$ and $\tau_{\text{mission\_small}} = 40\%$.

All campaigns are implemented in Python using the PyTorch framework and the PyTorch Image Models (\texttt{timm}) library~\cite{wightman2019timm}. The campaigns are executed on an AMD Ryzen 5 7600X CPU with process-level parallelism managed via \texttt{torch.set\_num\_threads}.

To maximize throughput across tens of thousands of experimental trials, two software optimizations are implemented:
\begin{enumerate}
    \item \textbf{In-Memory Tensor Preloading:} The validation subsets are pre-loaded directly into RAM as static tensors prior to entering the injection loop, removing disk-read overhead from the inference path.
    \item \textbf{Single-Parameter State Restoration:} Instead of performing a complete state-dictionary reload ($\Theta \leftarrow \text{load\_state\_dict}$) between consecutive trials, the runtime engine targets and swaps only the single corrupted parameter tensor using the memory address logged during the fault injection phase.
\end{enumerate}
Under the unoptimized flow (reloading the entire state dictionary), a single ViT-Tiny trial requires approximately $10$ seconds, culminating in a total campaign duration of $46.6$ hours for $16,538$ injections. Applying the single-parameter state restoration and tensor pre-loading optimizations to the larger ViT-Small network reduces the per-trial latency to $1.9$ seconds, allowing a campaign of $16,575$ injections to complete in $9$ hours. For reproducibility, random number generators in Python, NumPy, and PyTorch are initialized with a static seed.

\subsection{Experimental Results}
\label{res:main}

The experimental campaigns are structured to resolve vulnerabilities at increasing levels of specificity: first establishing the global failure probability, then localizing risk within architectural components, and finally isolating the numerical sensitivity of the FP32 representation. Each stage utilizes the empirical failure rate from the preceding level to optimize the sampling budget.

\subsubsection{Network-Wise Baseline (ViT-Tiny/EuroSAT)}
\label{sec:res_baseline}

The baseline campaign characterizes the global susceptibility of the network to random bit-flips across the entire parameter space. A total of $16,538$ trials were conducted, sampling $0.30\%$ of the $5,526,346$ trainable parameters. Table~\ref{tab:baseline} summarizes the global metrics.

\begin{table}[h]
\centering
\caption{Network-wise baseline results for ViT-tiny/EuroSAT ($e = 1\%$, 99\% confidence, $p = 0.5$).}
\label{tab:baseline}
\begin{tabular}{lr}
\hline
\textbf{Metric} & \textbf{Value} \\
\hline
Sample size ($n$) & 16,538 \\
Population fraction & 0.30\% \\
Failure rate ($\hat{p}$) & 2.93\% \\
Margin of error ($\hat{e}$) & $\pm 0.34\%$ \\
99\% CI & [2.60\%, 3.27\%] \\
Duration & 46.6 hours \\
\hline
\end{tabular}
\end{table}

The campaign yields an empirical failure rate ($\hat{p}$) of $2.93\%$, establishing a tight $99\%$ confidence interval of $[2.60\%, 3.27\%]$. These findings show that approximately $97.07\%$ of single-bit flips in the model's FP32 parameters do not cause operational failure. However, when failures do occur, they are heavily weighted toward catastrophic classification states. A comprehensive breakdown of these distinct failure categories is presented in Table~\ref{tab:failure_types}.

\begin{table}[t]
\centering
\caption{Failure type breakdown for the ViT-tiny/EuroSAT baseline campaign.}
\label{tab:failure_types}
\begin{tabular}{lrr}
\hline
\textbf{Failure type} & \textbf{Count} & \textbf{\% of total} \\
\hline
Catastrophic (NaN/Inf) & 98 & 0.59\% \\
Catastrophic (low acc) & 352 & 2.13\% \\
Mission failure & 15 & 0.09\% \\
SDC significant & 20 & 0.12\% \\
SDC minor & 234 & 1.41\% \\
No effect & 15,819 & 95.65\% \\
\hline
\end{tabular}
\end{table}

The observed failure behavior is highly polarized. Among the 485 critical failures falling into categories 1--4, 450 correspond to catastrophic outcomes, i.e., 92.8\% of all failures. Only 15 cases fall into the mission-failure range, while 20 produce significant but non-catastrophic silent data corruption. The region between no-effect behavior and severe collapse is therefore sparsely populated, indicating that a single bit-flip tends either to leave the model unaffected or to disrupt its functionality severely.

The validity of the normal approximation underlying Equation~\ref{eq:margin} is also satisfied. The standard rule of thumb requires $n \cdot \hat{p} \geq 5$ and $n \cdot (1 - \hat{p}) \geq 5$ for the binomial distribution to be adequately approximated by a normal distribution; here $n \cdot \hat{p} = 485 \gg 5$, so the condition holds comfortably.

\subsubsection{Layer-wise analysis}
\label{sec:res_layerwise}

\begin{figure}[t]
\centering
\includegraphics[width=\columnwidth]{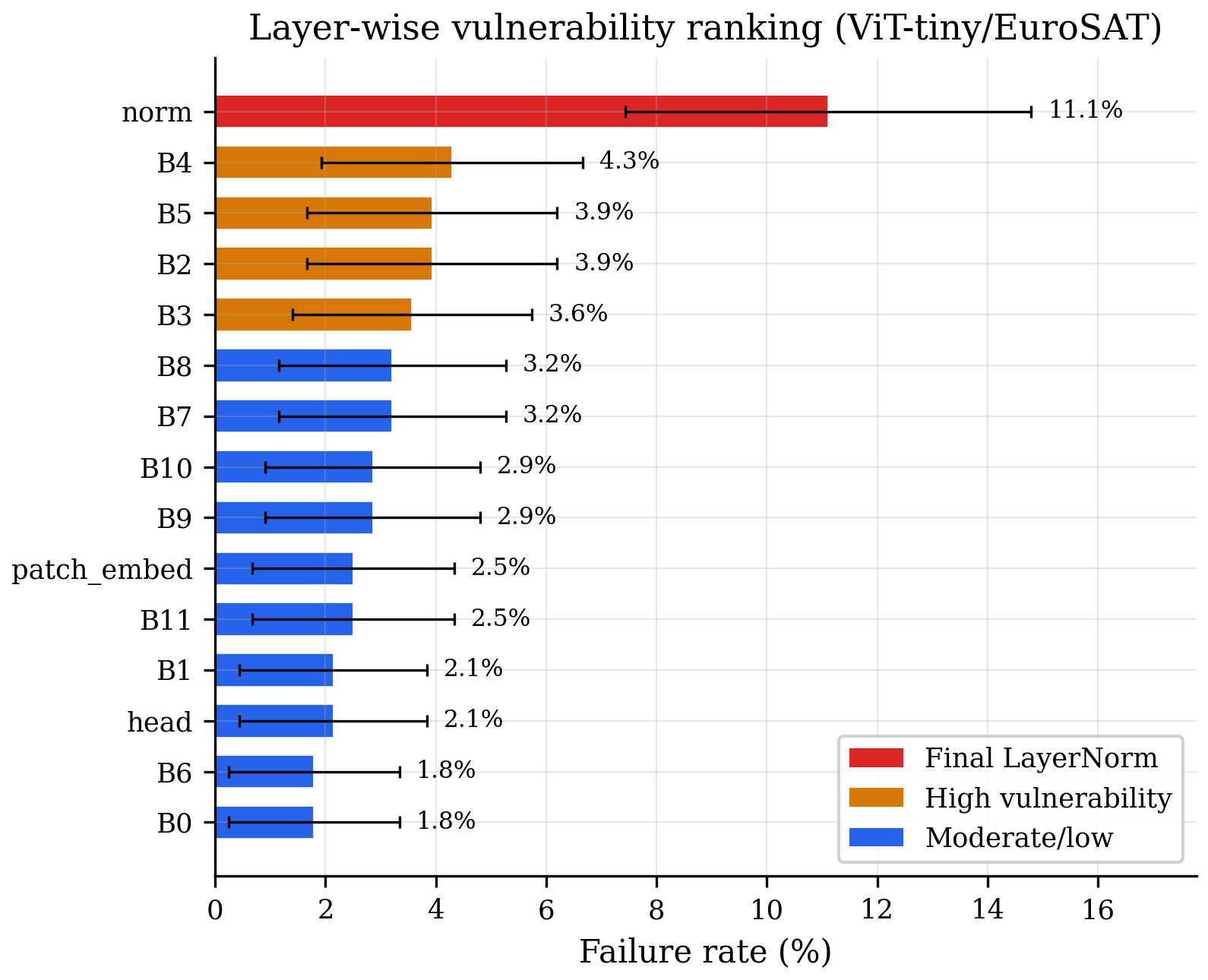}
\caption{Layer-wise failure rates with 95\% confidence interval error bars. The final LayerNorm (red) is a clear outlier at 11.1\%, while mid-layers (blocks 2--5, orange) cluster around 3.5--4.3\%. Most blocks overlap in their confidence intervals, but the LayerNorm is statistically separable from all of them.}
\label{fig:layerwise}
\end{figure}

Using the measured baseline estimate $\hat{p} = 0.03$ as prior information, separate SFI campaigns are conducted for each of the 15 layer-level populations. With $e = 2\%$ and 95\% confidence, the required sample size is approximately 280 tests per layer, giving a total of approximately $280 \times 15 = 4{,}200$ injections across all 15 layer populations. The complete layer-wise campaign requires 7.7 hours. Table~\ref{tab:layerwise} lists the most and least vulnerable layers, while Figure~\ref{fig:layerwise} provides the full ranking together with confidence interval error bars.

\begin{table}[t]
\centering
\caption{Layer-wise vulnerability ranking (ViT-tiny/EuroSAT, $e = 2\%$, 95\% confidence). Top 5 and bottom 3 layers shown.}
\label{tab:layerwise}
\begin{tabular}{clrrr}
\hline
\textbf{Rank} & \textbf{Layer} & \textbf{Failure rate} & \textbf{95\% CI} & \textbf{Parameters} \\
\hline
1 & norm & 11.11\% & [7.43\%, 14.80\%] & 384 \\
2 & block\_4 & 4.29\% & [1.91\%, 6.66\%] & 444,864 \\
3 & block\_2 & 3.93\% & [1.65\%, 6.20\%] & 444,864 \\
4 & block\_5 & 3.93\% & [1.65\%, 6.20\%] & 444,864 \\
5 & block\_3 & 3.57\% & [1.40\%, 5.74\%] & 444,864 \\
\hline
13 & block\_1 & 2.14\% & [0.47\%, 3.82\%] & 444,864 \\
14 & block\_0 & 1.79\% & [0.24\%, 3.34\%] & 444,864 \\
15 & block\_6 & 1.79\% & [0.24\%, 3.34\%] & 444,864 \\
\hline
\end{tabular}
\end{table}

The final LayerNorm (\texttt{norm}) emerges as the most vulnerable layer in Figure~\ref{fig:layerwise}. Its failure rate of 11.11\% is 3--5$\times$ higher than that of any transformer block, despite containing only 384 parameters, corresponding to 0.007\% of the model. Put differently, 0.007\% of the model's parameters account for a layer-level failure rate that exceeds every transformer block by a wide margin. This concentration of risk traces to the role of LayerNorm parameters as multiplicative scaling factors applied to every token in the sequence. A fault affecting one such scale parameter perturbs every activation passing through that dimension, and when this occurs in the final LayerNorm immediately before the classification head, there are no subsequent layers available to absorb or attenuate the error.

Among the transformer blocks, the highest failure rates are observed in the middle portion of the network, particularly blocks 2--5, whereas earlier blocks (0--1) and later blocks (6--11) show lower vulnerability. The confidence intervals in Figure~\ref{fig:layerwise} indicate that several neighboring blocks remain statistically overlapping; therefore, small differences in rank should not be over-interpreted. Nevertheless, the separation between the mid-layer group and the earlier/later layers appears more stable. One plausible explanation is that middle layers operate closer to a representational bottleneck, where errors are less easily diluted than in earlier or later stages benefiting from greater redundancy through residual paths.

\subsubsection{Component-wise analysis}
\label{sec:res_componentwise}

\begin{figure}[t]
\centering
\includegraphics[width=\columnwidth]{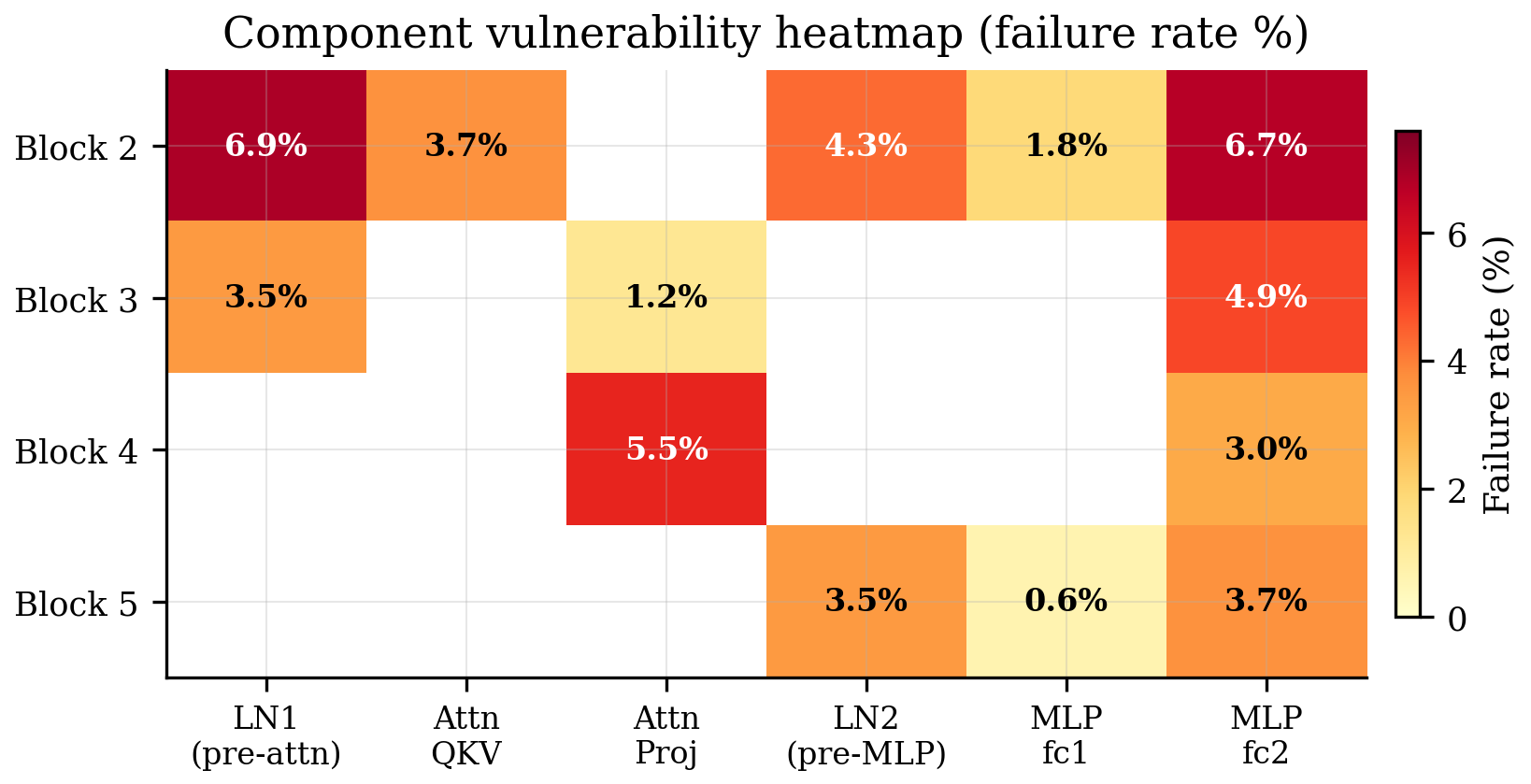}
\caption{Component vulnerability heatmap for blocks 2--5. Each cell shows the failure rate for a specific component type within a specific block. The left column (LN1) and right column (MLP fc2) are consistently darker, indicating higher vulnerability. The MLP fc1 column is consistently light.}
\label{fig:heatmap}
\end{figure}

The component-wise analysis further refines the vulnerability study by decomposing the most critical regions, namely the final LayerNorm layer and transformer blocks 2--5, into their constituent components. These include Attn~QKV, Attn~Proj, MLP~fc1, MLP~fc2, LN1 (pre-attention LayerNorm), and LN2 (pre-MLP LayerNorm). Table~\ref{tab:componentwise} reports the ten most vulnerable components, while Figure~\ref{fig:heatmap} presents the same information in heatmap form to highlight cross-block trends. The full campaign covers 25 components, with 116--164 tests per component, and requires 8.7 hours. The sample size for each component is computed independently using Equation~\ref{eq:sample_size}, with $N$ set to the number of parameters in that component and $p = 0.04$ as the prior. Because component populations are small (ranging from a few hundred to tens of thousands of parameters), the finite population correction in Equation~\ref{eq:margin} is non-negligible for several components, and this is what produces the variation in required tests across the 116--164 range.

\begin{table}[t]
\centering
\caption{Component vulnerability ranking, top 10 (ViT-tiny/EuroSAT, $e = 3\%$, 95\% confidence). Component labels correspond to those used in Figure~\ref{fig:heatmap}.}
\label{tab:componentwise}
\begin{tabular}{clrr}
\hline
\textbf{Rank} & \textbf{Component} & \textbf{Failure rate} & \textbf{95\% CI} \\
\hline
1 & Final LN & 8.62\% & [4.35\%, 12.89\%] \\
2 & B2 / LN1 & 6.90\% & [3.04\%, 10.75\%] \\
3 & B2 / MLP fc2 & 6.71\% & [2.88\%, 10.53\%] \\
4 & B4 / Attn Proj & 5.49\% & [2.01\%, 8.97\%] \\
5 & B3 / MLP fc2 & 4.88\% & [1.58\%, 8.17\%] \\
6 & B2 / LN2 & 4.31\% & [1.22\%, 7.40\%] \\
7 & B2 / Attn QKV & 3.66\% & [0.79\%, 6.53\%] \\
8 & B5 / MLP fc2 & 3.66\% & [0.79\%, 6.53\%] \\
9 & B3 / LN1 & 3.45\% & [0.67\%, 6.23\%] \\
10 & B5 / LN2 & 3.45\% & [0.67\%, 6.23\%] \\
\hline
\end{tabular}
\end{table}

Three main trends can be identified from these results, two of which are immediately visible in the heatmap of Figure~\ref{fig:heatmap}.

First, LayerNorm-related components dominate the highest ranks. The Final~LN component is ranked first, while B2/LN1, B2/LN2, and B3/LN1 also appear among the top entries. In the heatmap, the LN1 column remains among the most vulnerable across all analyzed blocks. This mirrors the layer-wise results and adds further evidence that normalization parameters carry outsized vulnerability.

Second, the second MLP layer (MLP~fc2) is consistently more vulnerable than the first (MLP~fc1). This contrast is visible across all analyzed blocks in Figure~\ref{fig:heatmap}, where the MLP~fc2 column appears systematically darker than MLP~fc1. In block 2, for example, MLP~fc2 exhibits a failure rate of 6.71\%, whereas MLP~fc1 shows only 1.83\%. Similar differences are observed in blocks 3 and 5. Architecturally, this behavior can be explained by the fact that MLP~fc2 writes directly back into the residual stream, so faults affecting this layer are injected immediately into the main representational path. By contrast, MLP~fc1 projects activations into a higher-dimensional hidden space, where perturbations may be partially attenuated by the subsequent GELU activation and the MLP~fc2 projection.

Third, the vulnerability of the attention output projection (Attn~Proj) varies across blocks. In block 4 it ranks fourth with a failure rate of 5.49\%, whereas in block 3 it reaches only 1.22\%. In general, the query-key-value projection (Attn~QKV) tends to be less vulnerable than the output projection (Attn~Proj). The softmax operation constrains attention coefficients to $[0,1]$, which limits how far perturbations originating in $Q$, $K$, or $V$ can propagate.

\subsubsection{Bit-wise analysis}
\label{sec:res_bitwise}
\begin{figure*}[h!]
\centering
\includegraphics[width=1.5\columnwidth]{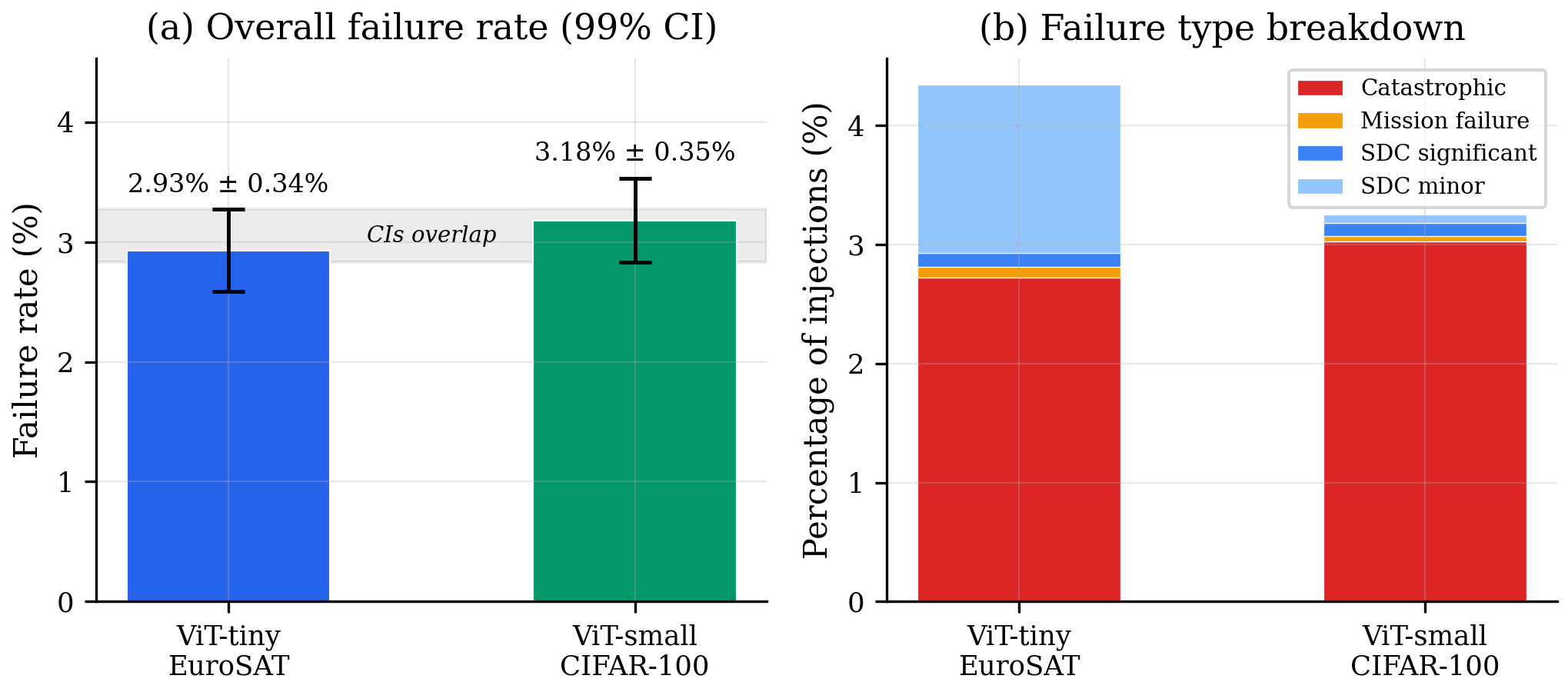}
\caption{Cross-architecture comparison. (a) Overall failure rates with 99\% confidence intervals. The gray band marks the region where the two CIs overlap, indicating no statistically significant difference. (b) Failure type breakdown showing the same catastrophic-dominant pattern in both configurations.}
\label{fig:cross_arch}
\end{figure*}
The bit-wise analysis quantifies the contribution of each FP32 bit position to the overall failure rate. Unlike the layer- and component-wise campaigns, which sample across all parameters within a population, each bit-wise campaign fixes one bit position and samples that position across \emph{all} trainable parameters in the model, i.e., the fault space for bit $k$ is the set of all weights with their $k$-th bit flipped. The sample size for each bit position is computed from Equation~\ref{eq:sample_size} using adaptive priors: $p = 0.001$ for mantissa bits (0--22) and the sign bit (31), giving $n \approx 5$ tests per bit position; $p = 0.05$ for exponent bits (23--29), giving $n \approx 200$ tests; and $p = 0.86$ for bit~30, giving $n \approx 514$ tests at $e = 3\%$ and 95\% confidence. Summing across all 32 bit positions yields approximately $32 \times \bar{n} \approx 2{,}000$ total injections. Table~\ref{tab:bitwise} reports the aggregated results by bit field. The complete campaign requires 4 hours of runtime.

\begin{table}[t]
\centering
\caption{Bit-wise failure rates aggregated by FP32 field (ViT-tiny/EuroSAT).}
\label{tab:bitwise}
\begin{tabular}{lrrrr}
\hline
\textbf{Bit range} & \textbf{Tests} & \textbf{Failures} & \textbf{Rate} & \textbf{Type} \\
\hline
Mantissa (0--22) & 115 & 0 & 0.00\% & 23 mantissa bits \\
Exponent (23--29) & 1,421 & 8 & 0.56\% & 7 exponent bits \\
Bit 30 (exp MSB) & 514 & 449 & 87.35\% & 1 exponent bit \\
Sign (bit 31) & 5 & 0 & 0.00\% & sign bit \\
\hline
\end{tabular}
\end{table}


A single bit position clearly dominates the failure behavior. Bit~30 accounts for 449 out of 457 observed failures within the bit-wise sub-campaign, i.e., 98.2\% of failures observed in that campaign. Referencing back to the network-wide baseline, bit~30 is responsible for approximately 95-96\% of all failures observed across the full fault space, and exhibits a failure rate of 87.35\% ($\pm 2.87\%$ at 95\% confidence). In IEEE-754 arithmetic, flipping the most significant exponent bit shifts the exponent by $\pm 128$, which changes the weight magnitude by a factor of up to $2^{128} \approx 3.4 \times 10^{38}$. Perturbations of this scale routinely cause overflow or NaN propagation.

By contrast, mantissa bits and the sign bit produce no observed failures in the ViT-Tiny campaign. However, these results should be interpreted cautiously because the per-bit sample sizes for these positions are small. The remaining exponent bits (23--29) exhibit a low but non-zero aggregate failure rate of 0.56\%, with failure probability increasing toward more significant exponent positions.

\subsubsection{Cross-architecture validation}
\label{sec:res_validation}

The ViT-Small/CIFAR-100 validation campaign is used to assess whether the observed patterns are specific to the ViT-Tiny/EuroSAT configuration or remain consistent across a larger model and a more complex dataset. Table~\ref{tab:cross_arch} compares the main baseline and bit-wise results for the two configurations, while Figure~\ref{fig:cross_arch} presents the overall failure rates and failure-type distributions.

\begin{table}[t]
\centering
\caption{Cross-architecture comparison of baseline and bit-wise results.}
\label{tab:cross_arch}
\small
\setlength{\tabcolsep}{3pt}
\begin{tabularx}{\columnwidth}{@{}l >{\centering\arraybackslash}X >{\centering\arraybackslash}X@{}}
\hline
\textbf{Metric} & \textbf{ViT-tiny/EuroSAT} & \textbf{ViT-small/CIFAR-100} \\
\hline
Parameters ($N$) & 5,526,346 & 21,704,164 \\
Sample size ($n$) & 16,538 & 16,575 \\
Failure rate & 2.93\% $\pm$ 0.34\% & 3.18\% $\pm$ 0.35\% \\
99\% CI & [2.60\%, 3.27\%] & [2.83\%, 3.53\%] \\
No effect & 95.65\% & 96.73\% \\
Cata. (of failures) & 92.8\% & 95.3\% \\
Duration & 46.6 h & 9.0 h \\
\hline
Bit 30 failure rate & 87.35\% $\pm$ 2.87\% & 87.73\% $\pm$ 2.84\% \\
Bit 30 tests & 514 & 888 \\
Bit 30 share of fail. & ${\sim}$96\% & 95.6\% \\
Mantissa (0--22) & 0/115 (0\%) & 0/4,600 (0\%) \\
Sign (31) & 0/5 & 0/200 \\
\hline
\end{tabularx}
\end{table}

Table~\ref{tab:cross_arch} shows that the two configurations produce closely matching baseline failure probabilities: 2.93\% for ViT-Tiny and 3.18\% for ViT-Small. Their 99\% confidence intervals overlap at [2.83\%, 3.27\%], indicating no statistically significant difference. The bit~30 failure rates are likewise nearly identical (87.35\% vs.\ 87.73\%), and mantissa bits remain failure-free under substantially larger sample sizes in ViT-Small (0/4,600 tests), confirming the ViT-Tiny finding. Figure~\ref{fig:cross_arch} presents the same results visually. Panel (a) makes the CI overlap explicit through the gray shaded band, and panel (b) shows that the catastrophic-dominant failure pattern holds in both configurations.

\begin{figure*}[th!]
\centering
\includegraphics[width=1.6\columnwidth]{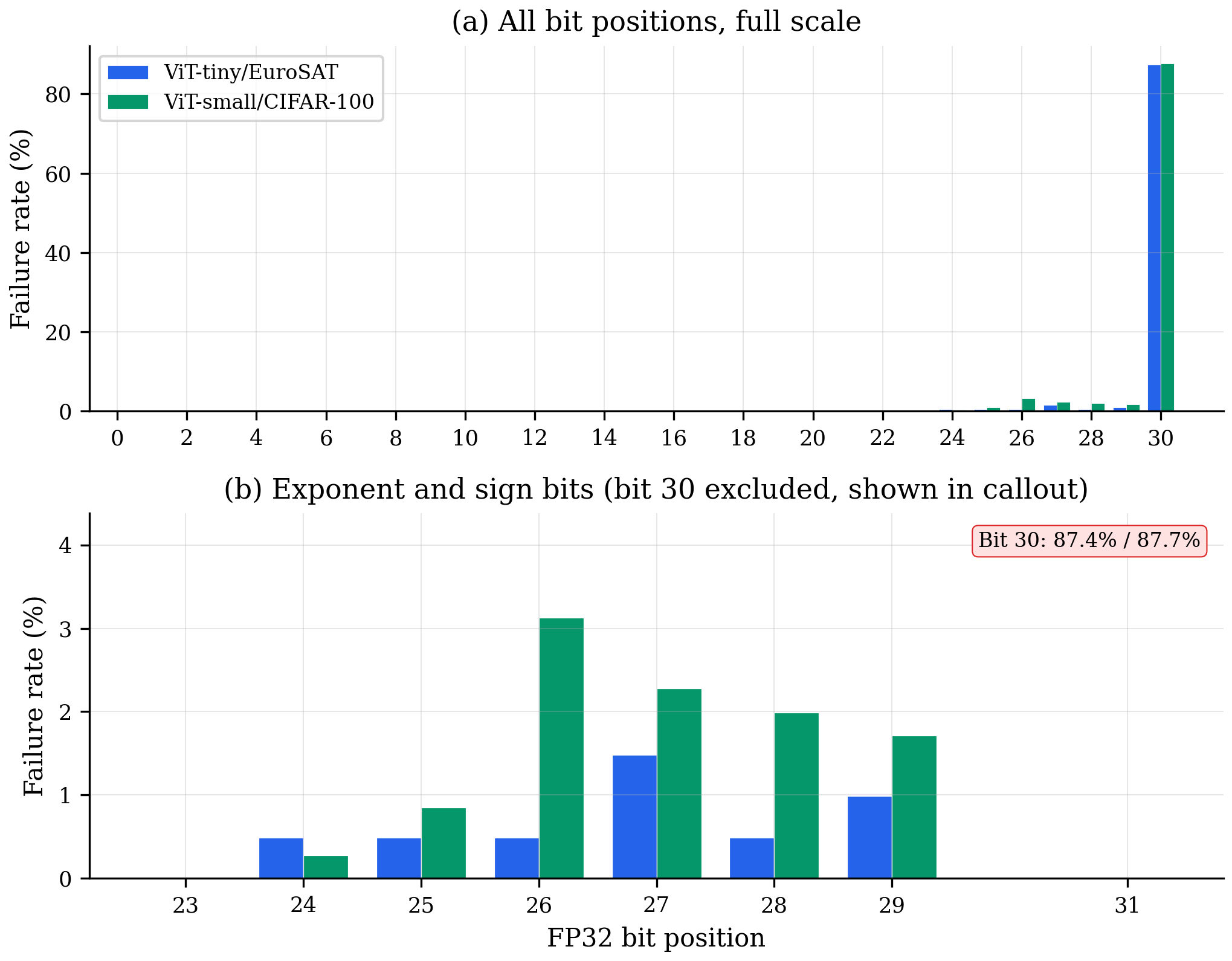}
\caption{Bit-wise failure rate comparison across architectures. (a) Full scale, showing matching bit-30 spikes at 87.4\% and 87.7\% for ViT-Tiny and ViT-Small, respectively. (b) Zoomed view of exponent bits 23--29 and sign bit 31 (bit 30 excluded for readability)}
\label{fig:bitwise_comparison}
\end{figure*}

Figure~\ref{fig:bitwise_comparison} compares the per-bit profiles directly.

The higher sampling budget in the ViT-Small campaign makes the non-dominant exponent-bit behavior more interpretable. With 351 tests per exponent bit at 99\% confidence, a monotonic gradient becomes visible across bits 23--29, with higher-order exponent bits producing higher failure probabilities. Bit~26 exhibits the largest non-bit-30 failure rate at 3.13\% (CI [0.74\%, 5.53\%]), while bits 27--29 range from 1.71\% to 2.28\%. Bit~23 shows no observed failures in 351 tests.

The validation campaign also resolves the main statistical limitation of the ViT-Tiny bit-wise experiment. Mantissa bits remain failure-free in ViT-Small as well, but now under substantially larger sample sizes (0/4,600 total tests), yielding meaningful confidence bounds instead of the weak per-bit estimates obtained in the original ViT-Tiny campaign.

\subsubsection{Speedup analysis}
\label{sec:res_speedup}

\begin{table}[t]
\centering
\caption{Fault injection cost comparison for ViT-Tiny/EuroSAT. All SFI timings use the same hardware (CPU) and the same per-injection evaluation on 1,000 test images. Exhaustive estimates assume the same 10~s/injection runtime.}
\label{tab:speedup}
\small
\begin{tabularx}{\columnwidth}{@{}Xrr@{}}
\hline
\textbf{Approach} & \textbf{Tests} & \textbf{Est. time} \\
\hline
Exhaustive (all params $\times$ 32 bits) & 176,843,072 & ${\sim}$56 years \\
Exhaustive (all params $\times$ 1 bit) & 5,526,346 & ${\sim}$1.75 years \\
\hline
SFI baseline only ($p{=}0.5$, $e{=}1\%$, 99\%) & 16,538 & 46.6 h \\
SFI all granularity levels combined & ${\sim}$26,000 & ${\sim}$67 h \\
\hline
\end{tabularx}
\end{table}

Table~\ref{tab:speedup} summarizes the computational savings achieved by the proposed statistical methodology. At the network level, the baseline SFI campaign reduces the experimental effort from 176.8 million injections to 16,538, corresponding to a speedup of approximately 10,700$\times$, while still providing formal 99\% confidence bounds with $\pm 0.34\%$ margin of error. Reusing the measured failure probability $\hat{p} \approx 0.03$ for subsequent campaigns further reduces the required sample size at layer granularity from roughly 16,500 to about 280 tests per layer.

Across all six campaigns considered in this study, namely baseline, layer-wise, component-wise, and bit-wise analysis on ViT-Tiny together with baseline and bit-wise validation on ViT-Small, the total effort amounts to approximately 51,000 injection experiments completed in about 80 hours. In contrast, exhaustive bit-level evaluation of ViT-Tiny alone would require 176 million injections. Because the SFI sample size grows only weakly with population size (Equation~\ref{eq:sample_size}), the relative speedup grows with model scale. Across all granularity levels on ViT-Tiny, the full multi-granularity campaign requires roughly 26,000 injections compared to 176 million for exhaustive coverage, a reduction of approximately $6{,}800\times$.


\section{Conclusion}
\label{sec:conclusion}

This work introduces a statistically grounded fault injection framework that replaces heuristic testing with formal confidence bounds. Through iterative multi-granularity sampling, the proposed methodology achieves a $10,700\times$ efficiency gain, enabling reliable ViT vulnerability assessment without the prohibitive costs of exhaustive evaluation. The empirical results reveal a paradox that while $97\%$ of single-bit flips are benign, the remaining $3\%$ are disproportionately catastrophic. These failures are not random; they concentrate heavily on Bit 30 of the FP32 exponent and critical LayerNorm parameters.  This work provides a practical basis for hardware-software co-design, with concrete targets for lightweight hardening such as selective ECC for exponent bits and activation-clamping for normalization layers.

\section*{Acknowledgements}

This work was supported in part by the Estonian Research Council grant PUT PRG1467 "CRASHLESS“, EU Grant Project 101160182 “TAICHIP“, by the Deutsche Forschungsgemeinschaft (DFG, German Research Foundation) – Project-ID "458578717", and by the Federal Ministry of Research, Technology and Space of Germany (BMFTR) for supporting Edge-Cloud AI for DIstributed Sensing and COmputing (AI-DISCO) project (Project-ID "16ME1127").


\bibliographystyle{IEEEtran}
\bibliography{ref}

@inproceedings{leveugle2009statistical,
  author    = {Leveugle, R. and Calvez, A. and Maistri, P. and Vanhauwaert, P.},
  title     = {Statistical Fault Injection: Quantified Error and Confidence},
  booktitle = {Design, Automation and Test in Europe Conference (DATE)},
  year      = {2009},
  pages     = {502--506},
  doi       = {10.1109/DATE.2009.5090716},
  note      = {TIMA Laboratory, Grenoble INP, UJF, CNRS}
}

@article{ruospo2025quantitative,
  author    = {Ruospo, Annachiara and Sonza Reorda, Matteo and Mariani, Riccardo and Sanchez, Ernesto},
  title     = {An Effective Iterative Statistical Fault Injection Methodology
               for Deep Neural Networks},
  journal   = {IEEE Transactions on Computers},
  volume    = {74},
  pages     = {2431--2444},
  year      = {2025},
  issn      = {0018-9340},
  doi       = {10.1109/TC.2025.3566863}
}

@article{xue2023soft,
  author    = {Xue, Xinghua and Liu, Cheng and Wang, Ying and Yang, Bing and
               Luo, Tao and Zhang, Lei and Li, Huawei and Li, Xiaowei},
  title     = {Soft Error Reliability Analysis of Vision Transformers},
  journal   = {IEEE Transactions on Very Large Scale Integration (VLSI) Systems},
  volume    = {31},
  number    = {12},
  pages     = {2126--2136},
  year      = {2023},
  doi       = {10.1109/TVLSI.2023.3317138}
}

@article{he2025fine,
  author    = {He, Jiajun and Liu, Yi and Xu, Changqing and Liao, Xinfang and Yang, Yintang},
  title     = {Fine-Grained Fault Sensitivity Analysis of Vision Transformers
               Under Soft Errors},
  journal   = {Electronics},
  volume    = {14},
  pages     = {2418},
  year      = {2025},
  doi       = {10.3390/electronics14122418}
}

@inproceedings{liao2025analyzing,
  author    = {Liao, En-Yu and Wang, Ting-Chi},
  title     = {Analyzing and Enhancing the Reliability of Vision Transformer
               Models Against Soft Errors},
  booktitle = {IEEE International Symposium on Circuits and Systems (ISCAS)},
  year      = {2025},
  pages     = {1--5},
  doi       = {10.1109/ISCAS58744.2025.11043840}
}

@inproceedings{dosovitskiy2021image,
  author    = {Dosovitskiy, Alexey and Beyer, Lucas and Kolesnikov, Alexander and
               Weissenborn, Dirk and Zhai, Xiaohua and Unterthiner, Thomas and
               Dehghani, Mostafa and Minderer, Matthias and Heigold, Georg and
               Gelly, Sylvain and Uszkoreit, Jakob and Houlsby, Neil},
  title     = {An Image is Worth 16$\times$16 Words: Transformers for Image
               Recognition at Scale},
  booktitle = {International Conference on Learning Representations (ICLR)},
  year      = {2021}
}

@article{baumann2005radiation,
  author    = {Baumann, R.},
  title     = {Radiation-Induced Soft Errors in Advanced Semiconductor Technologies},
  journal   = {IEEE Transactions on Device and Materials Reliability},
  volume    = {5},
  number    = {3},
  pages     = {305--316},
  year      = {2005},
  doi       = {10.1109/TDMR.2005.852615}
}

@article{helber2019eurosat,
  author    = {Helber, Patrick and Bischke, Benjamin and Dengel, Andreas and Borth, Damian},
  title     = {{EuroSAT}: A Novel Dataset and Deep Learning Benchmark for Land Use
               and Land Cover Classification},
  journal   = {IEEE Journal of Selected Topics in Applied Earth Observations
               and Remote Sensing},
  volume    = {12},
  number    = {7},
  pages     = {2217--2226},
  year      = {2019},
  doi       = {10.1109/JSTARS.2019.2918242}
}

@inproceedings{reagen2018ares,
  author    = {Reagen, Brandon and Gupta, Udit and Pentecost, Lillian and
               Whatmough, Paul and Lee, Sae Kyu and Mulholland, Niamh and
               Brooks, David and Wei, Gu-Yeon},
  title     = {Ares: A Framework for Quantifying the Resilience of Deep Neural Networks},
  booktitle = {ACM/IEEE Design Automation Conference (DAC)},
  pages     = {1--6},
  year      = {2018},
  doi       = {10.1145/3195970.3195997}
}

@misc{wightman2019timm,
  author    = {Wightman, Ross},
  title     = {{PyTorch Image Models}},
  year      = {2019},
  howpublished = {GitHub},
  url       = {https://github.com/huggingface/pytorch-image-models},
  doi       = {10.5281/zenodo.4414861}
}

@article{hendrycks2016gaussian,
  title={Gaussian error linear units ({GELU}s)},
  author={Hendrycks, Dan and Gimpel, Kevin},
  journal={arXiv preprint arXiv:1606.08415},
  year={2016}
}

@inproceedings{1,
  title={DMR-based Technique for Fault Tolerant AES S-box Architecture},
  author={Taheri, Mahdi and Sheikhpour, Saeideh and Ansari, Mohammad Saeed and Mahani, Ali},
  booktitle={1 st Conference on Applied Research in Electrical Engineering (AREE)},
  year={2021},
  organization={http://aree1.ir/papers/AREE1-ELTC-08.pdf}
}

@article{2,
  title={A fault-resistant architecture for aes s-box architecture},
  author={Taheri, Mahdi and Sheikhpour, Saeideh and Ansari, Mohammad Saeed and Mahani, Ali},
  journal={Journal of Applied Research in Electrical Engineering},
  volume={1},
  number={1},
  pages={86--92},
  year={2022},
  publisher={Shahid Chamran University of Ahvaz}
}

@inproceedings{3,
  title={A novel fault-tolerant logic style with self-checking capability},
  author={Taheri, Mahdi and Sheikhpour, Saeideh and Mahani, Ali and Jenihhin, Maksim},
  booktitle={2022 IEEE 28th International Symposium on On-Line Testing and Robust System Design (IOLTS)},
  pages={1--6},
  year={2022},
  organization={IEEE}
}

@article{4,
  title={Noise-tolerance gpu-based age estimation using resnet-50},
  author={Taheri, Mahtab and Taheri, Mahdi and Hadjahmadi, Amirhossein},
  journal={arXiv preprint arXiv:2305.00848},
  year={2023}
}

@article{5,
  title={FORTUNE: A Negative Memory Overhead Hardware-Agnostic Fault TOleRance TechniqUe in DNNs},
  author={Nazari, Samira and Taheri, Mahdi and Azarpeyvand, Ali and Afsharchi, Mohsen and Ghasempouri, Tara and Herglotz, Christian and Daneshtalab, Masoud and Jenihhin, Maksim},
  journal={Authorea Preprints},
  year={2024},
  publisher={Authorea}
}

@inproceedings{6,
  title={Reliability-aware performance optimization of DNN HW accelerators through heterogeneous quantization},
  author={Nazari, Samira and Taheri, Mahdi and Azarpeyvand, Ali and Afsharchi, Mohsen and Herglotz, Christian and Jenihhin, Maksim},
  booktitle={2025 IEEE 26th Latin American Test Symposium (LATS)},
  pages={1--6},
  year={2025},
  organization={IEEE}
}

@inproceedings{7,
  title={Genie: Genetic algorithm-based reliability assessment methodology for deep neural networks},
  author={Nazari, Samira and Taheri, Mahdi and Azarpeyvand, Ali and Afsharchi, Mohsen and Herglotz, Christian and Jenihhin, Maksim},
  booktitle={2025 11th International Conference on Computing and Artificial Intelligence (ICCAI)},
  pages={264--271},
  year={2025},
  organization={IEEE}
}

@article{8,
  title={Reliability-Aware Hyperparameter Optimization for ANN-to-SNN Conversion},
  author={Sharifian, Saeed and Taheri, Mahdi and Rashtchi, Vahid and Azarpeyvand, Ali and Herglotz, Christian and Jenihhin, Maksim},
  journal={WiPiEC Journal-Works in Progress in Embedded Computing Journal},
  volume={11},
  number={1},
  pages={7--7},
  year={2025}
}

@inproceedings{9,
  title={RESQ: A Unified Framework for ReLiability-and SEcurity Enhancement of Quantized Deep Neural Networks},
  author={Mohammadi, Ali Soltan and Nazari, Samira and Azarpeyvand, Ali and Taheri, Mahdi and Krstic, Milos and H{\"u}bner, Michael and Herglotz, Christian and Ghasempouri, Tara},
  booktitle={2026 IEEE 27th Latin American Test Symposium (LATS)},
  pages={1--4},
  year={2026},
  organization={IEEE}
}

@article{10,
  title={Mix-and-Match Pruning: Globally Guided Layer-Wise Sparsification of DNNs},
  author={Monachan, Danial and Nazari, Samira and Taheri, Mahdi and Azarpeyvand, Ali and Krstic, Milos and Huebner, Michael and Herglotz, Christian},
  journal={arXiv preprint arXiv:2603.20280},
  year={2026}
}

@inproceedings{11,
  title={Adaptive Fault Resilience for Early-Exit DNNs},
  author={Kodamanchili, Rama Mounika and Cherezova, Natalia and Taheri, Mahdi and Jenihhin, Maksim},
  booktitle={2025 IEEE International Test Conference in Asia (ITC-Asia)},
  pages={108--113},
  year={2025},
  organization={IEEE}
}

@article{12,
  title={PhD Thesis Summary: Methods for Reliability Assessment and Enhancement of Deep Neural Network Hardware Accelerators},
  author={Taheri, Mahdi},
  journal={arXiv preprint arXiv:2603.08724},
  year={2026}
}

@techreport{krizhevsky2009learning,
  author      = {Krizhevsky, Alex},
  title       = {Learning Multiple Layers of Features from Tiny Images},
  institution = {University of Toronto},
  year        = {2009},
  url         = {https://www.cs.toronto.edu/~kriz/learning-features-2009-TR.pdf}
}

\end{document}